\documentclass[10pt,twocolumn,letterpaper]{article}

\usepackage{wacv}
\usepackage{times}
\usepackage{epsfig}
\usepackage{graphicx}
\usepackage{amsmath}
\usepackage{amssymb}
\usepackage{booktabs}
\usepackage{multirow}
\usepackage[accsupp]{axessibility}

\def\wacvPaperID{529} 

\wacvfinalcopy 

\ifwacvfinal
\usepackage[breaklinks=true,bookmarks=false]{hyperref}
\else
\usepackage[pagebackref=true,breaklinks=true,colorlinks,bookmarks=false]{hyperref}
\fi

\begin{document}

\title{Feature Disentanglement Learning with Switching and Aggregation \\for Video-based Person Re-Identification}

\author{
   Minjung Kim$^1$ \quad
   MyeongAh Cho$^1$ \quad
   Sangyoun Lee$^{1,2}$ \quad
   \vspace{0.01cm}\\
   $^1$ Yonsei University $^2$ Korea Institute of Science and Technology (KIST)\\
   {\tt\small \{mjkima, maycho0305, syleee\}@yonsei.ac.kr}

}

\maketitle
\thispagestyle{empty}

\begin{abstract}
In video person re-identification (Re-ID), the network must consistently extract features of the target person from successive frames. Existing methods tend to focus only on how to use temporal information, which often leads to networks being fooled by similar appearances and same backgrounds. In this paper, we propose a Disentanglement and Switching and Aggregation Network (DSANet), which segregates the features representing identity and features based on camera characteristics, and pays more attention to ID information. We also introduce an auxiliary task that utilizes a new pair of features created through switching and aggregation to increase the network's capability for various camera scenarios. Furthermore, we devise a Target Localization Module (TLM) that extracts robust features against a change in the position of the target according to the frame flow and a Frame Weight Generation (FWG) that reflects temporal information in the final representation. Various loss functions for disentanglement learning are designed so that each component of the network can cooperate while satisfactorily performing its own role. Quantitative and qualitative results from extensive experiments demonstrate the superiority of DSANet over state-of-the-art methods on three benchmark datasets.
\end{abstract}

\begin{figure}[t]
\centering
\includegraphics[width=1.0\columnwidth]{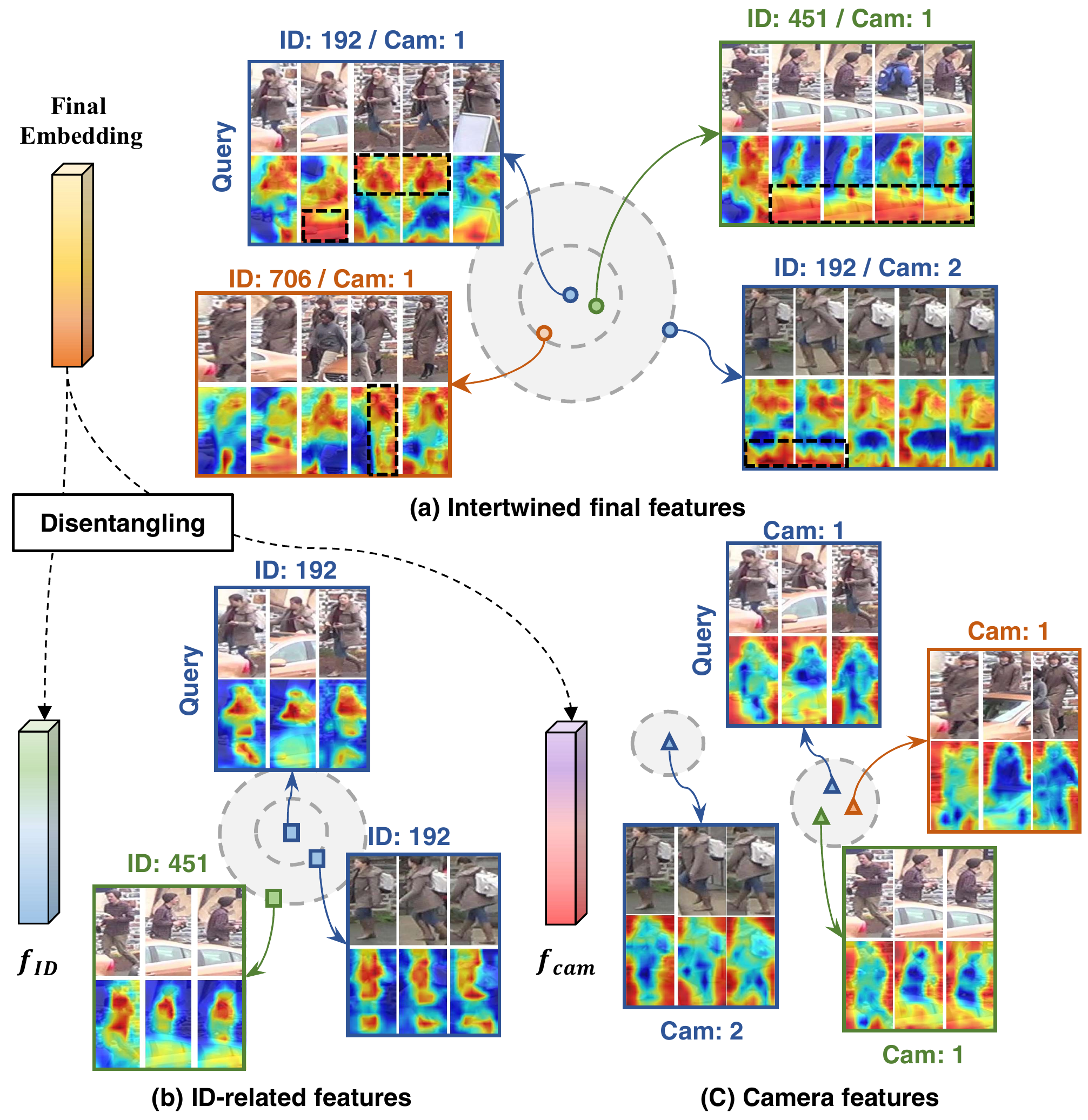} 
\caption{Challenges with consecutive frames as input in video Re-ID. (a) The final features are intertwined with background information and target-person information (especially dashed box).
Our proposed DSANet disentangles features into (b) ID-related features and (c) Camera features. Concentric circles represent distances in vector space around the query. The shapes of the embedding vectors denote different embedding spaces.}
\label{fig1}
\end{figure}

\section{Introduction}

Person re-identification(Re-ID)~\cite{ye2021deep} aims at matching the identity of a specific person on camera from various positions and angles. The demand for Re-ID has risen owing to the proliferation of intelligent surveillance systems and various multimedia applications. Most Re-ID approaches~\cite{li2021diverse, luo2019bag, zhang2020relation} employ deep metric learning to map embedding vectors corresponding to positive samples being closer to each other in vector space, and negative samples being farther away from each other. 

Unlike the self-supervised approaches~\cite{chen2020simple, wang2021residual} that compose positive samples through various data augmentation approaches, in Re-ID, positive samples comprise images having the same ID but belonging to different scenes. In other words, there are differences in characteristics such as background and obstacles depending on the camera taken, but if the ID is the same, the distance between embedding vectors should be low in vector space. Features extracted from the network inevitably include not only ID-related information but also camera domain/ID-unrelated information because the background and angle change depending on the scenes. This issue is more pertinent to video Re-ID, which uses several consecutive frames as input.

Compared to image Re-ID~\cite{he2020guided, he2019foreground}, video Re-ID provides temporal information which reduces the uncertainty of appearance. However, as ID-unrelated semantics also increase, the network may be vulnerable to intra-class noise. Most methodologies exploit temporal information such as obtaining between frames \cite{li2019global, wang2021pyramid}, selectively capturing details based on the previous frame features \cite{eom2021video, hou2021bicnet}, and extracting features that were not in the past sequence \cite{hou2020temporal}. However, these methods depend heavily on the network's capability extracting the correct features for the target during multiple frames.

For instance, as the time interval between frames is small, a similar background is repeated, so the network frequently learns the background as a key feature as in the activation map in Figure \ref{fig1}-(a), e.g. the dashed-box areas. As shown in Figure \ref{fig1}-(a), the final embedding includes the semantics of bricks and taxis, so cases with a similar appearance and background to the query are retrieved. In actual fact, unnecessary background information is repeated in the sequence and included in the final feature, which leads the network to determine that a vector of a person appearing in a scene having a similar appearance and background as the query is closer to the query than vectors of the same person appearing in a scene having a different appearance and background. Ultimately, if the network is forcing the final embedding vector with such entangled information to be closer to the vector space, the representational ability of the feature reduces.

In this paper, we propose a learning method that disentangles features and utilizes them to efficiently harness the network's capability without harming the representational ability. As briefly illustrated in Figure \ref{fig1}-(b, c), when the features are separated, the network fully focuses on each task, and the consistent rich information takes a pivot role to find the correct snippet as rank-1. Our proposed method, disentangling loss and camera ID classification, forces to separate features focused on identity from features having repeated background and occlusion patterns depending on the camera chosen. Furthermore, the target localization module (TLM) makes ID-related embeddings more robust against bounding box misalignment, a chronic concern in video tasks. By leveraging the disentangled features through switching and aggregation, the network learns to not only explicitly separate features but also implicitly increase capability. Our proposed method shows that it can be quite useful  for separation depending on the presence or absence of identity information in video Re-ID.

In summary, our main contributions are three-fold:
\begin{itemize}
\item We propose a novel DSANet that disentangles features into ID representative features and camera characteristic features and creates a new combination of these two for auxiliary tasks.
\item We introduce ID representation learning, which makes the final embedding features more robust to bounding box misalignment and discriminative to temporal flow.
\item Extensive experiments not only quantitatively demonstrate the excellent performance and competitive amount of parameters of our DSANet but also qualitatively prove disentanglement learning performs as we intended.
\end{itemize}

\begin{figure*}[t]
\centering
\includegraphics[width=1.0\textwidth]{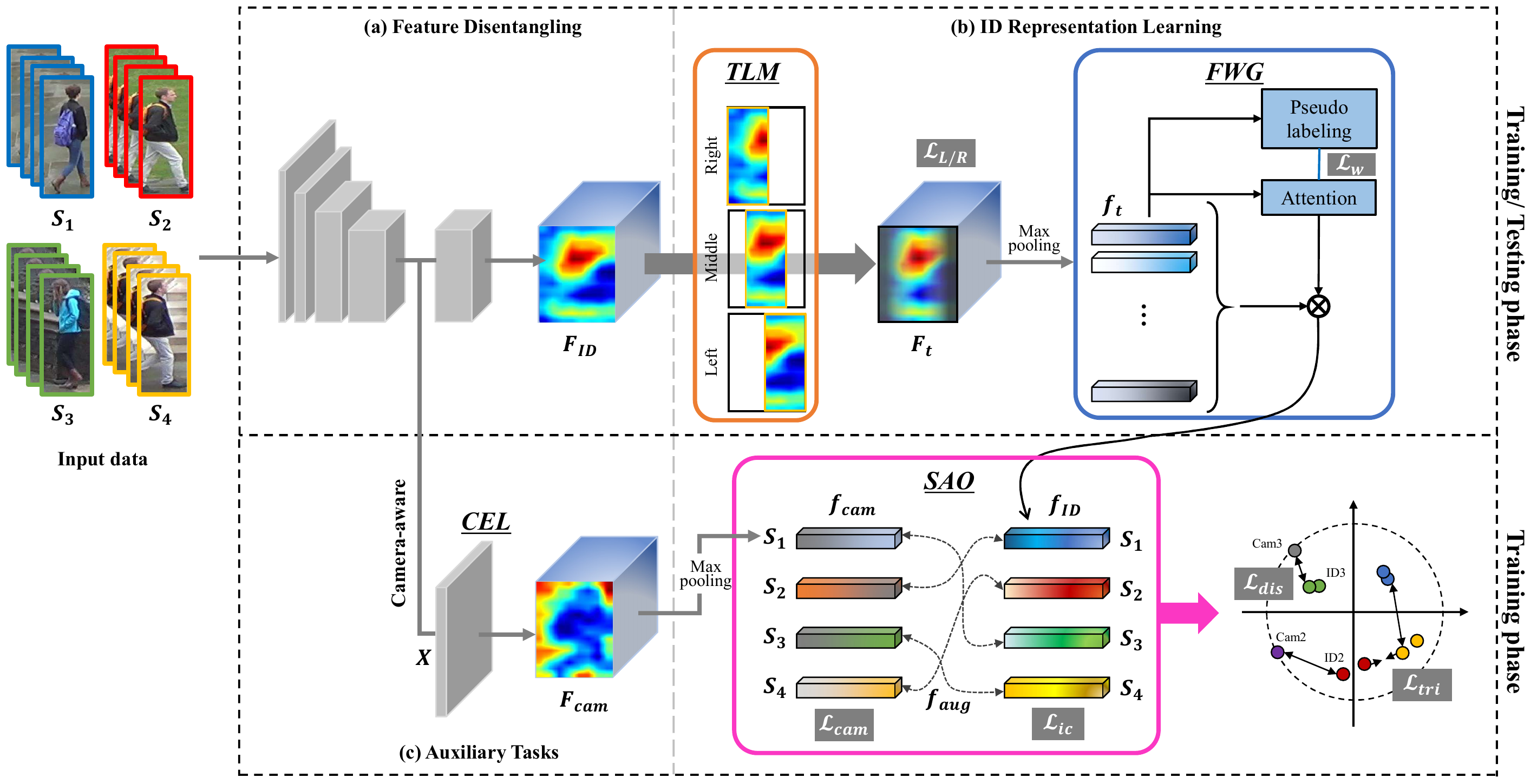}
\caption{Overall framework of our DSANet for video Re-ID. In the training phase, as input, the mini-batch consists of positive samples ($S_2$, $S_4$) and negative samples ($S_1$, $S_2$, $S_3$) based on ID. (a) \textbf{Feature Disentangling} process separates ID representative feature $\textbf{F}_{ID}$ and Camera Characteristic feature $\textbf{F}_{cam}$. In the (b) \textbf{ID representation learning}, TLM extracts consistent features for the target's position that changes over time, and FWG reflects temporal information in the final embedding. Then, SAO creates features with various camera scenarios and improves the DSANet's capacity through (c) \textbf{Auxiliary Tasks} that classify IDs. In the testing phase, we use $\textbf{f}_{ID}$ only as the final representation. White text in the gray box denotes loss function.}
\label{fig2}
\end{figure*}
\section{Related Works}
\subsection{Video-based Person Re-Identification}
In video Re-ID, it is crucial to completely extract the features of the target, excluding distractors. Most methods obtain robust video-level representations by utilizing the given information: spatial and temporal.
A \textbf{3D CNN}, capable of encoding the local temporal relation and the relative position, is used in conjunction with a non-local operation~\cite{gu2020appearance} or an attention mechanism~\cite{jiang2021ssn3d} that aligns each part to prevent deterioration of the final embeddings. Some studies~\cite{fu2019sta, li2019global} fully harness global-local \textbf{temporal clues} to predict the weights of each frame. Chen~\etal~\cite{chen2020temporal} focuses on the object, a time-invariant feature, rather than on extracting the motion vector. It disentangles features into temporal coherence and temporal motion and distills the discriminative characteristic by sampling noise at multiple scales. However, discriminative features are not necessarily immutable. For instance, depending on the video sequences, the background and obstacle may not change with a time dimension; rather, ID-related semantics may be variant because of the human pose. The most popular approach~\cite{eom2021video, hou2021bicnet, hou2019vrstc, wang2021pyramid, zhang2020multi} takes advantage of rich \textbf{spatio-temporal} clues. Hou~\etal~\cite{hou2021bicnet} considers temporal relations by adaptively selecting the temporal kernel scale and then extracts spatial features that are robust for multi-scaling. To make the final representation resistant to spatial and temporal distractors, Eom~\etal~\cite{eom2021video} identifies the distractor pattern, stores it in memory, and refines the video-level features through obstacle handling. These methods guide the network to extract rich representations of the target person but require a complex network structure and heavy computation. In light of \cite{eom2021video}'s observation that patterns occur in  backgrounds and obstacles, our method explicitly disentangles the features of the ID and background. Unlike \cite{chen2020temporal}, which obtains temporal coherence features by computing the mean, we propose an auxiliary task that can complement the disentangling process and enhance the discrimination of the network through feature augmentation. 

\subsection{Feature Disentanglement}
The feature disentangling method separates representations so that they have independent and intentional attributes and then utilizes them further. It is applied for diverse purposes~\cite{chu2021learning, huang2018multimodal, lee2021learning} in various areas, mainly in the domain adaptation field~\cite{lee2021dranet, zou2020joint}. Zheng~\etal~\cite{zheng2019joint} applies feature disentanglement in the Re-ID field. It has two distinct encoders to separate features based on appearance and structure code. The network generates high-quality cross-ID composed images by switching the codes and performs online learning with the  generated images. This series of processes goes on end-to-end. This method succeeds in reconstructing images characterized by mixed appearance and structure but fails to consider accessories and hair  corresponding to ID-related fine details in the appearance code. As mentioned in \cite{zhu2020identity}, the fine detail that imparts discriminant power is the crucial factor, but it remains in the structure code and may confuse network learning. Zou~\etal~\cite{zou2020joint} disentangles embeddings into ID-related/unrelated features and uses only ID-related features to reduce the huge domain gap. It has an encoder–decoder structure that reconstructs the original image using the cycle consistency characteristic to separate features into appearance and structure factors. Unlike the above methods, our method can satisfactorily disentangle features without requiring another network. In addition, switching and aggregation with disentangled vectors occurs at the feature level with no need to reconstruct the image.

\section{Proposed Method}
\subsection{DSANet}
A brief overview of our Disentanglement and Switching and Aggregation Network (DSANet) is given in Figure~\ref{fig2}. DSANet involves three processes: feature disentangling, ID representation learning, and auxiliary tasks. DSANet explicitly separates features into two categories: ID features and camera features. Then, ID representation learning is performed for the extracted ID  feature to become more resistant to changes in the target's position through Target Localization Module (TLM) and Frame Weight Generation (FWG). Finally, DSANet conducts auxiliary tasks by utilizing Switching and Aggregating Operation (SAO), which augments features to have various camera characteristics to improve the network's discernment. The series of processes learn end-to-end, and each component works individually and cooperatively to achieve the foremost goal of Re-ID: robust and differential feature extraction. The following sections present a detailed description of each part.
\subsection{Feature Disentangling}
\textbf{Channel Expansion Layer (CEL)}
Existing methods \cite{lee2021learning, zheng2019joint, zou2020joint} require multiple networks to disentangle features. However, DSANet separates $\textbf{F}_{ID}$ and $\textbf{F}_{cam}$ from one network satisfactorily by dividing the branches of the backbone. The starting point of dividing branches is the last layer of ResNet-50~\cite{he2016deep} before extracting more discriminative features. Relatively coarse $\textbf{X}\in \mathbb{R}^{\frac{c}{2}\times t\times h\times w}$ obtained from the third layer of ResNet passes through CEL consisting of $1\times 1$ convolution and returns $\textbf{F}_{cam}\in \mathbb{R}^{c\times t\times h\times w}$ containing semantics that specify the camera being considered. The last layer of the backbone returns $\textbf{F}_{ID}\in \mathbb{R}^{c\times t\times h\times w}$ containing only identity information. For $\textbf{F}_{cam}$ to separate reliably, we need the disentangling loss $\mathcal{L}_{dis}$ and the camera ID classification.

\textbf{Disentangling Loss}
Considering that $\textbf{f}_{ID}\in \mathbb{R}^{c\times 1 \times 1}$ and $\textbf{f}_{cam}\in \mathbb{R}^{c\times 1 \times 1}$ should have different information, we measure the cosine similarity of these two vectors as follows:
\begin{gather}
    \mathcal{L}_{dis} = max(\frac{\textbf{f}_{ID}\cdot \textbf{f}_{cam}}{\left \| \textbf{f}_{ID} \right \|_{2}{\left \| \textbf{f}_{cam} \right \|_{2}}}, 0),
\end{gather}
where $\textbf{f}_{cam}$ is vector maxpooling in the feature maps $\textbf{F}_{cam}$. When the cosine distance of $\textbf{f}_{ID}$ and $\textbf{f}_{cam}$ becomes 0, it indicates dissimilarity between the two vectors. A margin was given to the loss function because being exactly opposite could harm the representation ability. Disentangling loss allows DSANet to extract $\textbf{f}_{ID}$ and $\textbf{f}_{cam}$ focused on each task without being disturbed by spatial distractors in a situation where consecutive frames are input.

\begin{figure}[t]
\includegraphics[width=1.0\columnwidth]{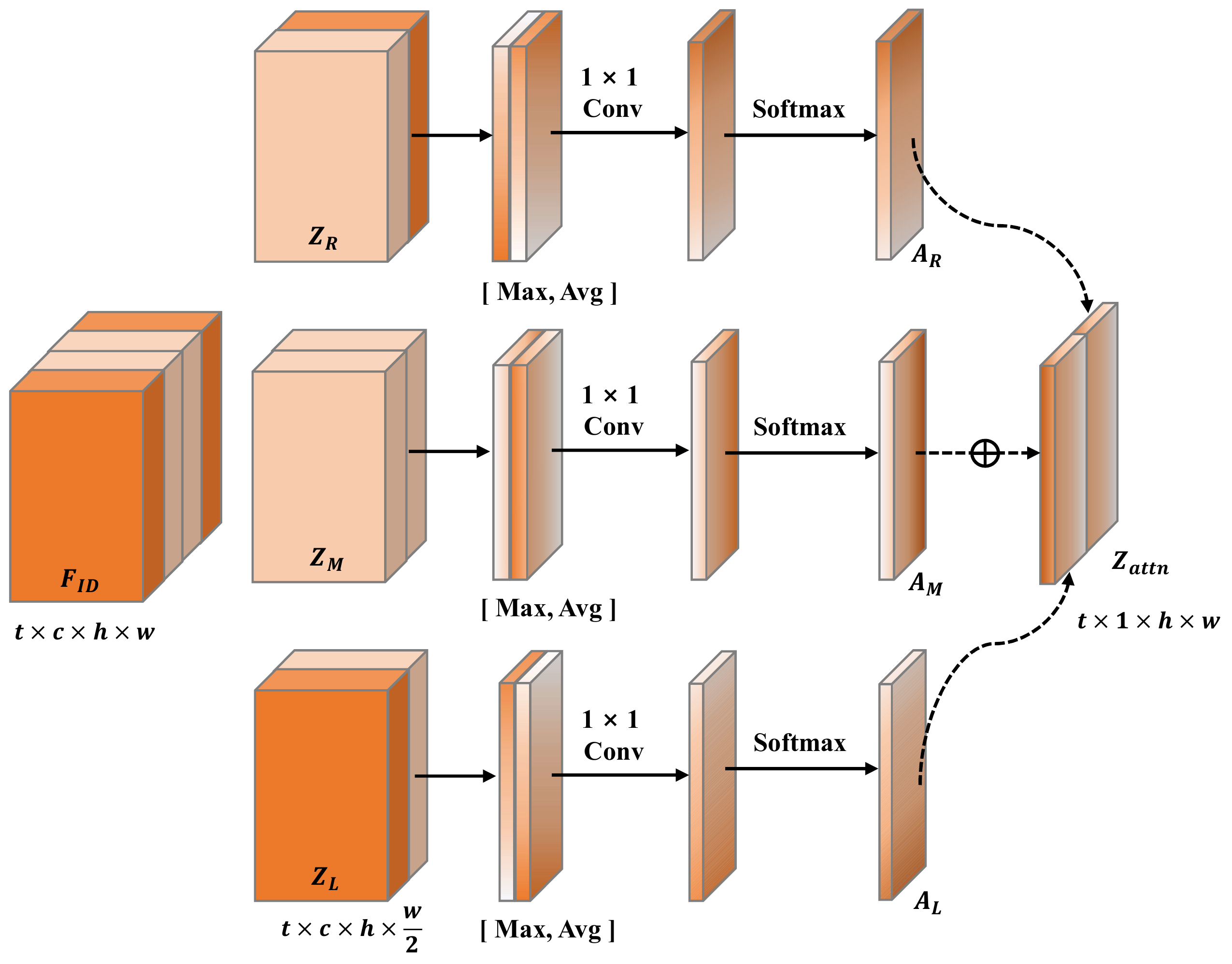} 
\caption{Illustration of Target Localization Module(TLM)}
\label{fig3}
\end{figure}

\subsection{ID representation Learning}
\textbf{Target Localization Module (TLM)}
To consistently extract $\textbf{F}_{ID}$ from the network even in consecutive frames, it must be robust to bounding box misalignment. One of the challenges in Re-ID is that the target is skewed in cropped video clips owing to the limits of the detector's performance. Therefore, in video Re-ID, the representation must follow the corresponding information according to the position of the moving person of interest. To achieve the above goal, we propose a TLM that conducts spatial attention, inspired by SAM~\cite{woo2018cbam}. It is designed to focus on where salient information is located in one complete feature map. However, our proposed TLM divides the feature map $\textbf{F}_{ID}$ into $\textbf{Z}_{L}$, $\textbf{Z}_{M}$, and $\textbf{Z}_{R}\in \mathbb{R}^{c\times t\times h\times \frac{w}{2}}$ in the spatial dimension as shown in Figure~\ref{fig3} and pays attention to the location of the target in each region.We concatenate the result of max pooling and average pooling in the channel dimension of the divided feature maps $\textbf{Z}$. Then we obtain spatial attention maps $A$ through a convolution layer and softmax.
\begin{gather}
    A = Softmax(Conv([Max(\textbf{Z});Avg(\textbf{Z})]))
\end{gather}
In many cases empirically, given that the person is located in the center~\cite{chi2020pedhunter, kim2022occluded}, we add information corresponding to the middle $A_{M}$ to make the final attention map $Z_{attn}$.
\begin{gather}
    Z_{attn} = [A_L; A_R] + A_M
\end{gather}
TLM finally acquires $\textbf{F}_t$ as the weighted sum of the $Z_{attn}$ and $\textbf{F}_{ID}$ so that it can flexibly cope with bounding box misalignment. 
\begin{gather}
    \textbf{F}_{t} = {Z}_{attn} \odot \textbf{F}_{ID} + \textbf{F}_{ID}
\end{gather}
TLM employs the following loss to consistently extract features by localizing the target in successive frames:
\begin{gather}
    \mathcal{L}_{L/R} = CE(\mathcal{P}(\textbf{f}_{t/L})) + CE(\mathcal{P}(\textbf{f}_{t/R})),
\end{gather}
where $CE$ stands for cross-entropy loss, $\mathcal{P}$ indicates the linear classifier computing probabilities, and $\textbf{f}_{t/L}\in \mathbb{R}^{c\times t\times 1\times 1}$ and $\textbf{f}_{t/R}\in \mathbb{R}^{c\times t\times 1\times 1}$ are vectors maxpooling in the feature maps corresponding to the left and right sides of $\textbf{F}_{t}$.

\begin{figure}[t]
\centering
\includegraphics[width=1.0\columnwidth]{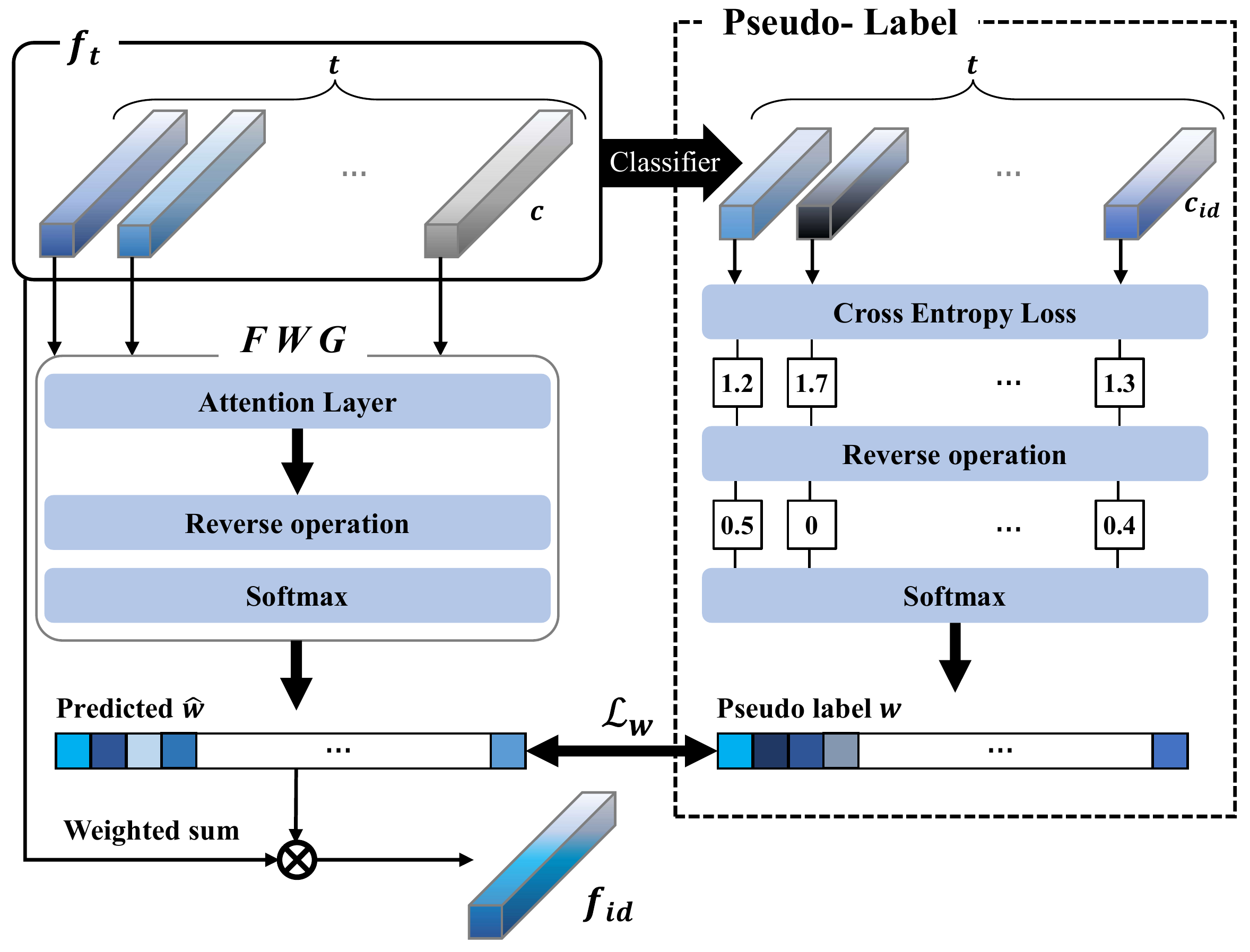} 
\caption{Illustration of Frame Weight Generation (FWG) and generating pseudo-label}
\label{fig4}
\end{figure}

\textbf{Frame Weight Generation (FWG)}
Most methods~\cite{mclaughlin2016recurrent, zhou2017see} mainly use global average pooling to merge the representation of the video sequence into the final embedding vector. To further utilize temporal information, we predict the weight $\hat{\textbf{w}}\in \mathbb{R}^{1\times t}$ for each frame through FWG and reflect this to obtain the final embedding vector. For the FWG to make reasonable predictions, we generate a pseudo label from the value that represents the importance index of the frame and use it for training. $\textbf{f}_t\in \mathbb{R}^{c\times t\times 1\times 1}$ was obtained by max-pooling of $\textbf{F}_t\in \mathbb{R}^{c\times t\times h\times w}$ extracted from TLM in the spatial dimension. $\textbf{f}_{t}$ contains a compact feature for each frame, which is input to the FWG and also used to generate a pseudo label. We compute the cross-entropy in the classification task, define it as a loss function, and learn to reduce this value. In other words, the smaller the cross-entropy value, the more correct the feature for the task. We can use this fact to determine which frame contains more identity information and express this value as a probability value. After dividing $\textbf{f}_{t}$ by frame, the cross-entropy with the ID label was computed respectively. Then, the max value was found among the obtained values, and the difference $d$ was calculated based on these values. We express this computation as a reverse operation in Figure~\ref{fig4}.
\begin{gather}
    d = max(CE(\mathcal{P}(\textbf{f}_t)))-CE(\mathcal{P}(\textbf{f}_t))
\end{gather}
We make these values' probabilities correspond to the importance of the frame and then use it as a pseudo-label $\textbf{w}$. 
FWG consists of one convolution layer reducing the channel dimension to 1. The weight is predicted by summarizing the frame features. The final weight $\hat{\textbf{w}}$ is predicted through the same operation process as pseudo label generation. We obtain the final representation $\textbf{f}_{ID}$ by taking temporal attention to $\hat{\textbf{w}}$ on $\textbf{f}_{t}$.

\textbf{Frame Weight Loss}
We calculate the mean squared error between pseudo label $\textbf{w}$ and predicted $\hat{\textbf{w}}$ and use it as a loss function $\mathcal{L}_{w}$ to expect reliable weights from the FWG.
 We do not use KL-divergence to reduce the difference between $\textbf{w}$ and $\hat{\textbf{w}}$ because the closer the probability value is to 0, the more difficult it is to measure the similarity between distributions. We also experimentally determine that the mean squared error is more appropriate for the reasonable prediction of $\hat{\textbf{w}}$.

\textbf{Intra-class Loss}
To reduce intra-class noise among the positive samples of the mini-batch, we define it as follows:
\begin{gather}
    \mathcal{L}_{ic} = \sum^{B}(\sum_{f_{ID}\in B_k}\frac{1}{B_k}(\textbf{f}_{ID}-\frac{1}{B_k}\sum_{f_{ID}\in B_k}\textbf{f}_{ID})^2), 
\end{gather}
where ${B}_{k}$ denotes the feature set belonging to identity $k$ in the mini-batch.
The $\textbf{f}_{ID}$ belonging to the positive samples is extracted from various sequences, but the semantic corresponding to the scene has been removed, so only the ID-related information is available. The intra-class loss function allows $\textbf{f}_{ID}$ to focus more on common but distinctive features within positive samples and reduces the burden on training.

\subsection{Auxiliary tasks}
\textbf{Camera ID Classification}
Although the dissimilarity of $\textbf{F}_{ID}$ and $\textbf{F}_{cam}$ is guaranteed through CEL and disentangling loss, a guide is needed for $\textbf{f}_{cam}$ to adaptively grasp the background and occlusion patterns depending on the camera. Figure~\ref{fig5} illustrates that the characteristics of the scene and obstacle patterns are similar depending on the camera taken for each dataset. Taking the above facts into account, DSANet performs the auxiliary task of predicting the camera ID with $\textbf{f}_{cam}$ containing the camera characteristics. As the camera IDs that captured the sequence are labeled, no additional annotation is needed. Rather, while actively utilizing the given label, the auxiliary task guides $\textbf{f}_{cam}$ to focus more on the background, excluding the target. The loss function used for camera ID classification is the cross-entropy loss calculated as follows:
\begin{gather}
    \mathcal{L}_{cam} = -c_{cam}^i\log(\frac{e^{\textbf{W}_i\textbf{f}^i_{cam}}}{\sum^{C_{cam}}_{j=1}e^{\textbf{W}_j\textbf{f}^j_{cam}}}),
    \label{CamCE}
\end{gather}
where $\textbf{W}$ is the weight matrix of the fully connected layer, $c_{cam}^i$ is the camera ID label, and $C_{cam}$ is the total number of camera IDs.

\textbf{Switching and Aggregating Operation (SAO)}
The disentangled $\textbf{f}_{ID}$ and $\textbf{f}_{cam}$ can be augmented using a new embedding vector through switching and aggregation. Specifically, when $\textbf{f}_{ID}$ and $\textbf{f}_{cam}$ constituting a mini-batch are randomly swapped and recombined, various pairs of $\textbf{f}_{ID}$ and $\textbf{f}_{cam}$ are generated. Even if a new $\textbf{f}_{cam}$ is assigned, the ID corresponding to the label of $\textbf{f}_{ID}$ does not change. Therefore, the ID of the embedding vector $\textbf{f}_{aug}$ obtained through SAO is determined by $\textbf{f}_{ID}$, so we use $\textbf{f}_{aug}$ for another auxiliary task of predicting the ID. This task not only improves the robustness of the network to various camera scenes but also enhances the disentangling learning adaptively.

In conclusion, the final cross-entropy loss used in DSANet is as follows:
\begin{gather}
    \mathcal{L}_{ce} = CE(\mathcal{P}(\textbf{f}_{ID})) + \lambda[CE(\mathcal{P}(\textbf{f}_{aug})) + \mathcal{L}_{L/R} + \mathcal{L}_{cam}].
\end{gather}

\subsection{Training and Testing Phases}
In the training phase, we employ the triplet loss~\cite{hermans2017defense} calculated with $\textbf{f}_{ID}$ and losses described in the previous section. The overall objective function of DSANet is defined as:
\begin{gather}
    \mathcal{L}_{total} = \mathcal{L}_{ce} + \mathcal{L}_{tri}+ \mathcal{L}_{dis} + \mathcal{L}_{ic} + \lambda \mathcal{L}_{w} , 
\end{gather}
where $\lambda$ is the scale factor and is set to 0.1.

In the testing phase, we use $\textbf{f}_{ID}$ only as the final representation and calculate the distance between the final embedding vectors using cosine similarity.

\begin{table*}[htbp]
  \tiny
  \centering
  \caption{Comparison with state-of-the-art methods on MARS, Duke-V, and LS-VID video Re-ID datasets. Methods are divided into three groups: 3D CNN-based, temporal-clues-based, spatio-temporal-clues-based. Best results in bold, second best underlined.}
  \vspace{3pt}
  	\resizebox{0.85
  	\linewidth}{!}{
				\renewcommand{\arraystretch}{1}
    \begin{tabular}{cc|r|cc|cc|cc}
    \hline
    \hline
    \multicolumn{2}{c|}{\multirow{2}*[-.3ex]{Methods}} & \multicolumn{1}{c|}{\multirow{2}*[-.3ex]{Params.}} & \multicolumn{2}{c|}{MARS~\cite{zheng2016mars}} & \multicolumn{2}{c|}{Duke-V~\cite{wu2018exploit}} & \multicolumn{2}{c}{LS-VID~\cite{li2019global}} \\
\cline{4-9}    \multicolumn{2}{c|}{} &       & \multicolumn{1}{c|}{Rank-1} & mAP   & \multicolumn{1}{c|}{Rank-1} & mAP   & \multicolumn{1}{c|}{Rank-1} & mAP \\
    \hline
    \multicolumn{2}{c|}{M3D~\cite{li2019multi} (AAAI2019)} &  \multicolumn{1}{c|}{-} & 84.4  & 74.1  &  -  & - & 57.7  & 40.1 \\
    \multicolumn{2}{c|}{AP3D~\cite{gu2020appearance} (ECCV2020)} & \multicolumn{1}{c|}{31.6 M} & 90.7  & 85.6  & \underline{97.2}  & 96.1  & -     & - \\
    \multicolumn{2}{c|}{SSN3D~\cite{jiang2021ssn3d} (AAAI2021)} & \multicolumn{1}{c|}{-} & 90.1  & 86.2  & 96.8  & 96.3  & -     & - \\
    \hline
    \multicolumn{2}{c|}{STA~\cite{fu2019sta} (AAAI2019)} & \multicolumn{1}{c|}{-} & 86.3  & 80.8  & 96.2  & 94.9  & -     & - \\
    \multicolumn{2}{c|}{GLTR~\cite{li2019global} (ICCV2019)} & \multicolumn{1}{c|}{-} & 87 & 78.5 & 96.3 & 93.7 & 63 & 44.3 \\
    \hline
    \multicolumn{2}{c|}{VRSTC~\cite{hou2019vrstc} (CVPR2019)} & \multicolumn{1}{c|}{-} & 88.5  & 82.3  & 95     & 93.5     & -     & -  \\
    \multicolumn{2}{c|}{MG-RAFA~\cite{zhang2020multi} (CVPR2020)} & \multicolumn{1}{c|}{-} & 88.8  & 85.9  & -     & -     & -     & -  \\
    \multicolumn{2}{c|}{TCLNet~\cite{hou2020temporal} (ECCV2020)} & \multicolumn{1}{c|}{29.9 M} & 89.8  & 85.1  & 96.9  & 96.2  & 81    & 67.2 \\
    \multicolumn{2}{c|}{AFA~\cite{chen2020temporal} (ECCV2020)} & \multicolumn{1}{c|}{-} & 90.2  & 82.9  & \underline{97.2}  & 95.4  & -     & - \\
    \multicolumn{2}{c|}{BiCnet-TKS~\cite{hou2021bicnet} (CVPR2021)} & \multicolumn{1}{c|}{29.2 M} & 90.2  & 86    & 96.1  & 96.3  & 84.6  & 75.1 \\
    \multicolumn{2}{c|}{STMN~\cite{eom2021video} (ICCV2021)} & \multicolumn{1}{c|}{-} & 90.5  & 84.5  & 97    & 95.9  & 82.1  & 69.2 \\
    \multicolumn{2}{c|}{PSTA~\cite{wang2021pyramid} (ICCV2021)} & \multicolumn{1}{c|}{35.4 M} & \textbf{91.5} & 85.8  & \textbf{98.3} & \textbf{97.4} & -     & - \\
    \hline
    \multicolumn{2}{c|}{DSANet (\textit{ours})} & \multicolumn{1}{c|}{30.8 M}  & \underline{91.1} & \textbf{86.6} & \underline{97.2} & \underline{96.6} & \textbf{85.1} & \textbf{75.5} \\
    \hline
    \hline
    \end{tabular}
  \label{t1}
  }
\end{table*}

\section{Experiments}

\subsection{Datasets and Evaluation Metrics}
\label{4.1}
\begin{figure}[t]
\centering
\includegraphics[width=0.9\columnwidth]{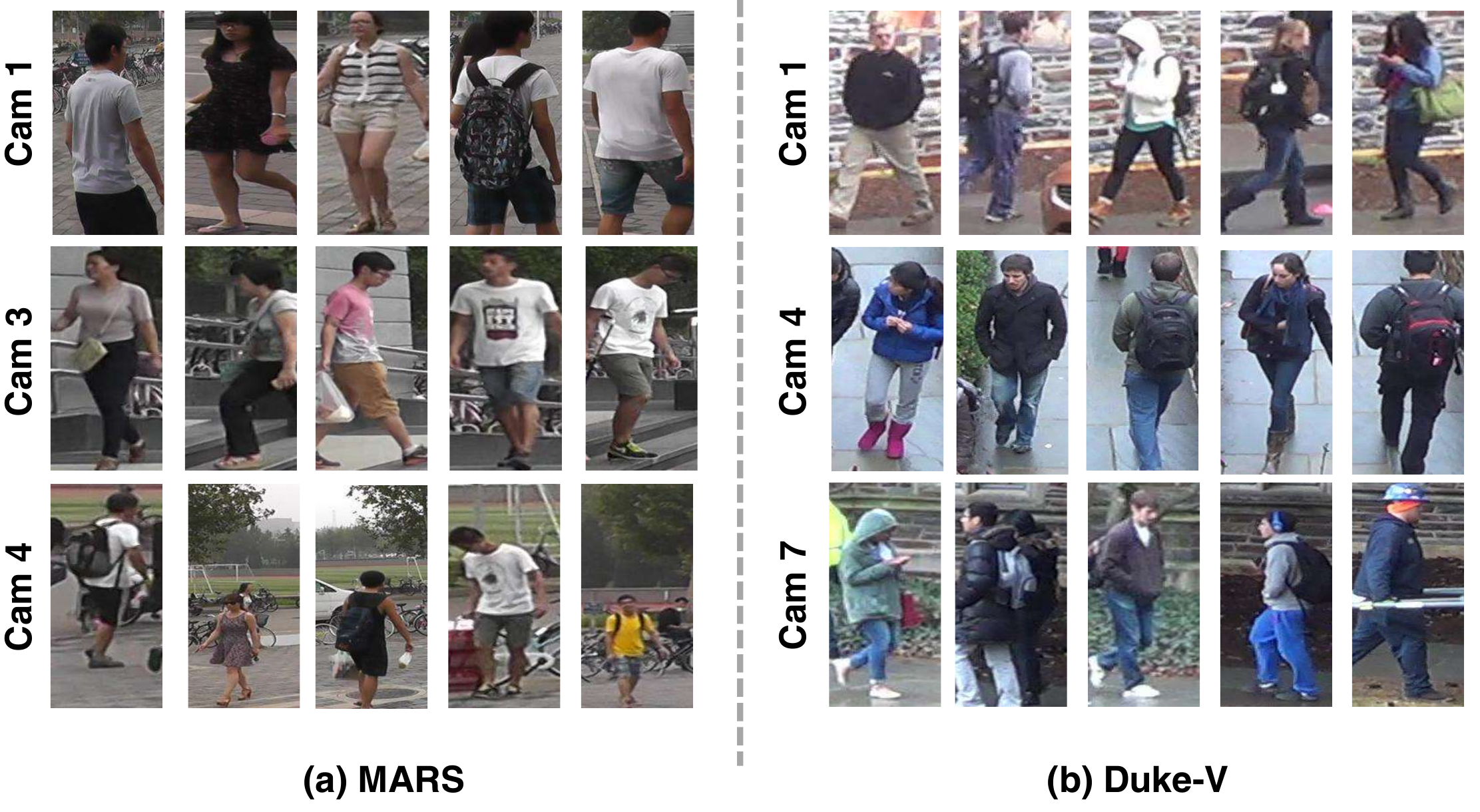} 
\caption{Cases of scene and obstacle patterns determined by the camera captured for each dataset (e.g. bricks, stairs, bicycles)}
\label{fig5}
\end{figure}

As shown in Figure~\ref{fig5}, characteristics of the scene and obstacle patterns are similar depending on the camera chosen for each dataset. Furthermore, in the Re-ID task, people's IDs do not overlap during the training and testing phase, but the domain of the chosen camera is the same regardless of the phase, so our method can be widely applied even if the number of cameras increases.

\textbf{MARS}~\cite{zheng2016mars} is a large-scale benchmark dataset for video Re-ID. It consists of approximately 20,000 tracklets of 1261 identities and additional distractors of 3248 tracklets. The video sequences are captured using 6 cameras. There are substantial bounding box misalignment problems, making it more realistic and challenging.

\textbf{DukeMTMC-Video ReID}~\cite{wu2018exploit} is another large-scale benchmark dataset, which contains 4,832 tracklets of 1,404 identities. The video sequences are captured using 8 cameras. We abbreviate DukeMTMC-VideoRe-ID as "Duke-V" in the following descriptions.

\textbf{LS-VID}~\cite{li2019global} is the most recent large-scale benchmark dataset for video Re-ID. It consists of 3,772 identities and 14,943 tracklets, captured using 15 cameras. There are many challenging elements such as illumination and bounding box misalignments, so it can be seen as the closest dataset to real life.

\textbf{Evaluation Metrics} The performance was evaluated using Cumulative Matching Characteristic (CMC) and mean Average Precision (mAP), which are frequently used as evaluation metrics in Re-ID.

\subsection{Implementation Details}
\label{4.2}
We adopt ResNet-50~\cite{he2016deep} pre-trained on ImageNet~\cite{deng2009imagenet} as a backbone of DSANet. To maintain the spatial dimension of the feature map, we change the stride of the last layer of ResNet-50 to 1. We randomly select 8 identities sampled with 4 clips for each identity to train model. We also leverage the restricted random sampling strategy~\cite{li2018diversity} to contain the entire video representation. Input frames are resized to 256×128, and data is augmented using random horizontal flip, and random erasing~\cite{zhong2020random} with a probability of 0.5. We use the Adam optimizer with a weight decay of $5\times10^{-4}$, whose learning rate starts at $3.5\times10^{-4}$ and decays by 0.1 times every 40 epoch. All experiments are conducted up to 200 epochs to ensure sufficient convergence of learning. In the testing phase, we compute all the frames of the sequence and obtain the final video-level feature through average pooling. As LS-VID has a large number of cameras, we set the lambda of $\mathcal{L}_{cam}$ to 0.5.

\begin{table}[t]
  \small
  \centering
  \caption{Analysis of each component (TLM, FWG, SAO) of DSANet including loss functions ($\mathcal{L}_{L/R}$, $\mathcal{L}_{w}$, $\mathcal{L}_{ic}$) on LS-VID.}
    \vspace{3pt}
    \begin{tabular}{ll|c|cc}
    \hline
    \hline
    \multicolumn{2}{c|}{\multirow{2}[2]{*}{Methods}} & \multirow{2}[2]{*}{Param.} & \multicolumn{2}{c}{LS-VID} \\
    \multicolumn{2}{c|}{} &       & Rank-1 & mAP \\
    \hline
    \multicolumn{2}{l|}{Baseline} & 25.2M & 73.3  & 62.1 \\
    \hline
    \multicolumn{2}{l|}{+TLM w/o $\mathcal{L}_{L/R}$} & 27.4M & 73.5  & 62.4 \\
    \multicolumn{2}{l|}{+TLM} & 29.1M & 73.7  & 62.5 \\
    \hline
    \multicolumn{2}{l|}{+FWG w/o $\mathcal{L}_{w}$} & 27.4M & 73.1  & 61.9 \\
    \multicolumn{2}{l|}{+FWG} & 29.1M & 74.1  & 62.5 \\
    \hline
    \multicolumn{2}{l|}{+TLM +FWG w/o $\mathcal{L}_{ic}$} & 30.8M & 73.9  & 62.5 \\
    \multicolumn{2}{l|}{+TLM +FWG} & 30.8M & 75.5  & \textbf{64.2} \\
    \hline
    \hline
    \multicolumn{2}{l|}{+TLM +FWG +SAO} & 30.8M & \textbf{75.8}    & \textbf{64.2} \\
    \hline
    \hline
    \end{tabular}
  \label{t2}
\end{table}

\subsection{Comparison with State-of-the-art Methods}
\label{4.3}
Before comparison, our network has a different model size for each dataset owing to the camera ID classifier. We list the size of our model for the LS-VID, which has the largest number of cameras in Table~\ref{t1}. The parameters of DSANet are 29.5M in MARS, and 30M in Duke-V.

Table~\ref{t1} summarizes the comparison between previous methods for three video Re-ID benchmark datasets. We divide the approaches into three groups: 3D-CNN-based, temporal-clues-based, spatio-temporal-clues-based. 

First, our method shows particularly good performance on MARS compared to 3D-CNN-based methods~\cite{gu2020appearance, jiang2021ssn3d, li2019multi}. MARS has frequent bounding box misalignment problems, so it is easy for the network to lose the consistency of temporal appearance when using 3D convolution. Our proposed method is robust against temporal appearance destruction because TLM aligns the position of the target that changes with the flow of the frame.

Compared to the methods using temporal clues~\cite{fu2019sta, li2019global}, our DSANet achieves much better performance for all datasets. Although temporal information is crucial in the video, approaches that utilize spatial information together can extract more discriminative features. As our method employs spatial information through TLM and obtains final embedding by reflecting temporal information through FWG, DSANet extracts a rich representative feature.

Finally, our method achieves superior or comparable (considering the number of parameters) performance to the spatio-temporal-clues-based methods~\cite{chen2020temporal, eom2021video, hou2021bicnet, hou2020temporal, hou2019vrstc,    wang2021pyramid, zhang2020multi}. DSANet uses not only spatio-temporal clues but also feature-level augmentation through SAO with ID features and camera features. It performs an auxiliary task that allows the network to classify IDs for various cases. While PSTA has significantly higher performance against Duke-V, the mAP of MARS close to real-life scenarios is 0.8\% lower than that of our method. PSTA~\cite{wang2021pyramid}, considering intra-frame and inter-frame relationships, is robust to long-term occlusion, but the inter-frame correlation operation and pyramid structure increase its parameters up to 35.4M. Our method obtains remarkable performance with a lightweight network without additional parameters through a simple feature-switching and aggregation mechanism. Performance can be further improved by combining other modules.

\textbf{Comparison with Related Methods}
We agree with the AFA~\cite{chen2020temporal} approach of separating features to focus on the target itself in video Re-ID. However, although AFA claims that the network should focus on the time-invariant characteristic of the target, targets are not the only ones that remain the same over time. Depending on the sequence, the background may change with time, and ID-related information that changes with time may rather be a discriminative feature. Furthermore, when compared with the network design, AFA tries to disentangle the features finally extracted from the backbone network.
However, our method extracts each feature from different branches with the goal of having dissimilar information in the feature stage. Furthermore, through visualization of the feature map corresponding to our camera feature, we can intuitively see which part each of the disentangled features has information about. Disentangled features are also utilized in SAO so that the network can extract robust and rich representations.

\begin{figure}[t]
\centering
\includegraphics[width=0.9\columnwidth]{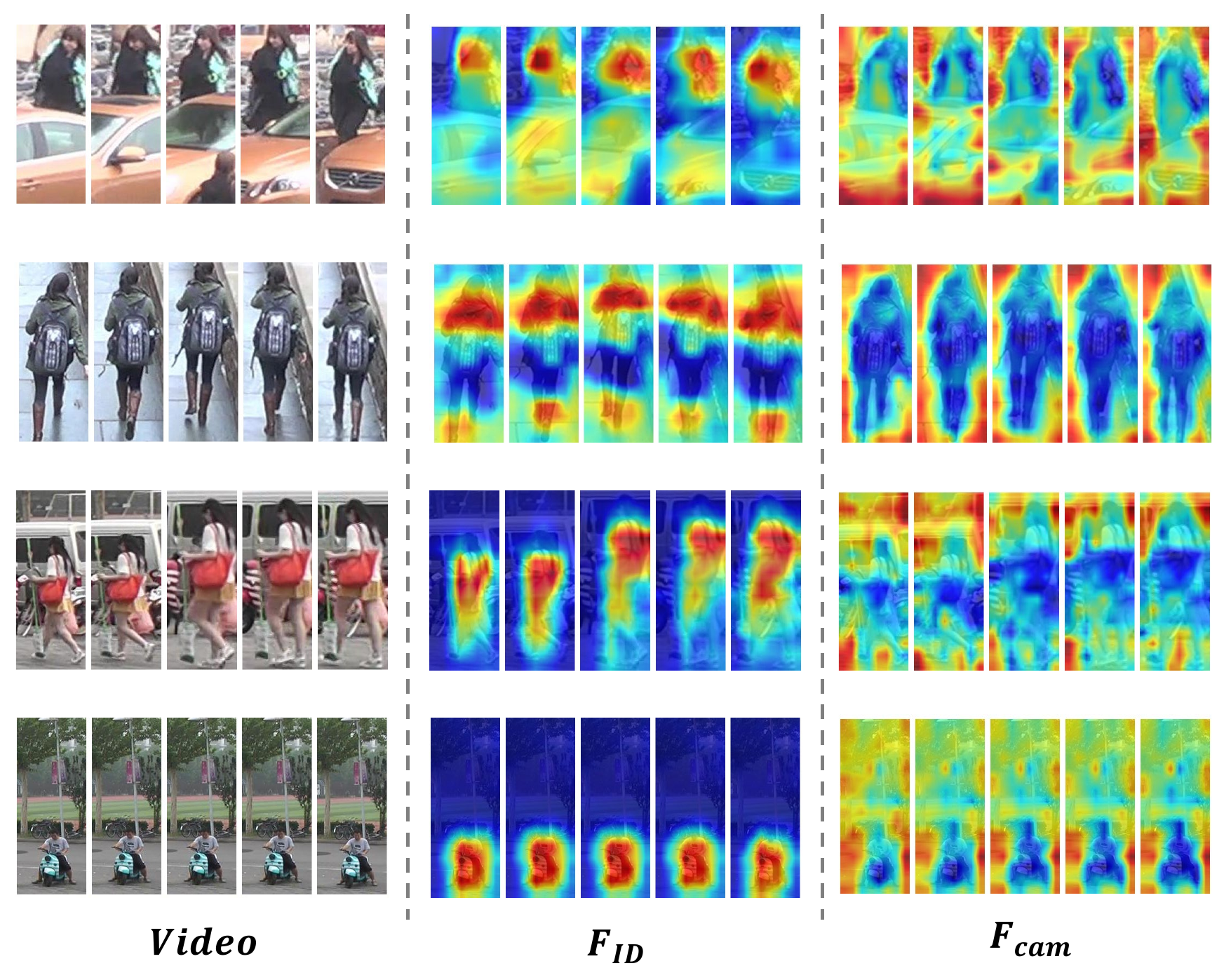}
\caption{Heatmap of disentangled features on MARS and Duke-V: $\textbf{F}_{ID}$ and $\textbf{F}_{cam}$. The warmer color denotes higher activation.}
\label{fig6}
\end{figure}

\subsection{Discussion}
\label{4.4}
\paragraph{Ablation Studies}
Table~\ref{t2} summarizes the results of ablation studies on the components (TLM, FWG, and loss functions) of DSANet. As the evaluation of the entire frame takes much time, we conduct the ablation studies according to the RRS~\cite{li2018diversity}. We also list the evaluation results for all frames in Table~\ref{t1}. If TLM or FWG is applied alone, it should be employed in conjunction with its loss function to aid network training. As TLM helps extract salient features for the target, it improves rank-1 by 0.4\% and mAP by 0.4\%. FWG also improves rank-1 by 0.8\% and mAP by 0.4\%. This is because FWG reflects temporal information in the final embedding to avoid interruption of the representation due to temporal obstructions. Finally, we can conclude that TLM and FWG are effective when used cooperatively, imporving  rank-1 by 2.2\% and mAP by 2.1\% from baseline. Furthermore, if the ID feature obtained from TLM and FWG is used for SAO, the network acquires the generalization ability to cover numerous situations. Surprisingly, despite SAO being an additional parameterless mechanism, it improves rank-1 by 0.3\%. In summary, DSANet's components work complementarily, so they are most effective when used together and perform best when all frames corresponding to the sequence are input.
\begin{figure}[t]
\centering
\includegraphics[width=1.0\columnwidth]{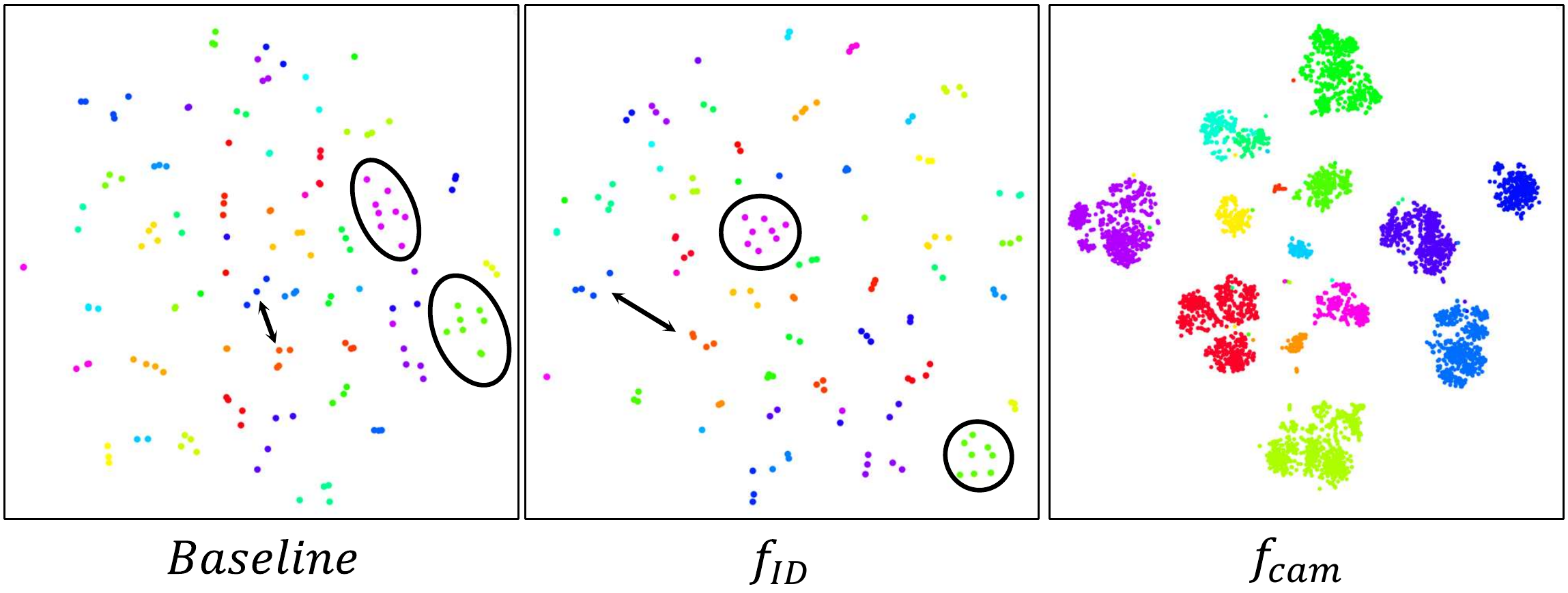} 
\caption{t-SNE~\cite{van2008visualizing} visualization of final embedding vectors from baseline, $\textbf{f}_{ID}$, and $\textbf{f}_{cam}$ from DSANet on LS-VID.}
\label{fig7}
\end{figure}

\textbf{Heatmap of Disentangled Features}
To prove we create and disentangle the features according to our proposed method, DSANet, we visualize the heatmap of each feature map in Figure~\ref{fig6}. In the first row corresponding to the case where obstacles continuously appear in the frame, DSANet obtains the final representation containing only information about the target. In the second and third rows, they show that DSANet can extract consistent features thanks to TLM despite the change of size and location of the target over time. As seen in the last row, even when the background occupies most of the frame, DSANet tries to extract discriminative identity information $F_{ID}$ by clearly separating the background information $F_{cam}$. In conclusion, DSANet successfully obtains final embeddings that are robust against background cluttering and bounding box misalignment problems while being rich in ID information, which is the primary goal in video Re-ID.

\textbf{Feature Distribution}
In Figure~\ref{fig7}, we visualize the disentangled features using t-SNE~\cite{van2008visualizing} to demonstrate that each feature contains either the identity or the camera information. We map the final embedding vector from baseline as well as $\textbf{f}_{id}$ and $\textbf{f}_{cam}$ from DSANet to the embedding space. Compared with the baseline, our DSANet can further narrow the intra-class distance through $\mathcal{L}_{ic}$. In addition, complementary learning of SAO, TLM, and FWG allows DSANet to extract distinguishing features of $\textbf{f}_{id}$ and widen the inter-class distance. Finally, we infer that the auxiliary task for predicting the camera ID is effective based on the results of grouping each other according to $\textbf{f}_{cam}$.

\begin{figure}
\centering
\includegraphics[width=0.8\columnwidth]{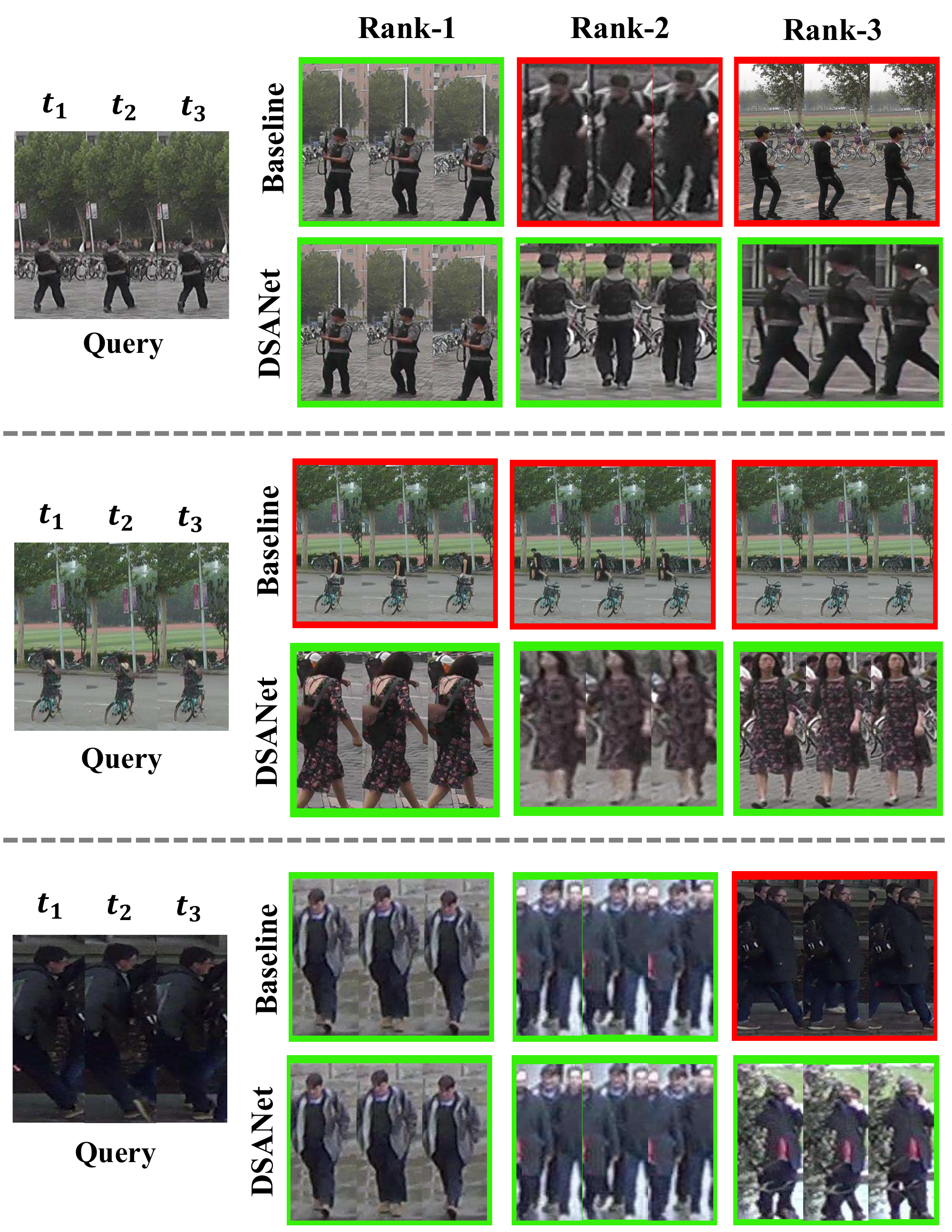}
\caption{Retrieval results of baseline and DSANet on MARS and Duke-V. The correct and incorrect matches are bordered in  green and red, respectively.}
\label{fig8}
\end{figure}

\textbf{Results of Retrieval}
Figure~\ref{fig8} shows the results of retrieval on MARS and Duke-V. In MARS, there are many samples with the background occupying more of the image than humans. This can be seen as the characteristic that can specify the camera that took the sequence. As the baseline cannot separate this camera characteristic, so background information is included in the final embedding, so a similar background to the query is retrieved. By contrast, DSANet can completely focus on identity information because the characteristics of the captured camera are perfectly disentangled. Even for the last row sample, the baseline was distracted by the background, so information that could specify the identity is missed, and the correct answer is not found until rank-10. Finally, DSANet shows excellent results of retrieval by focusing on fine details without being disturbed.

\section{Conclusion}
We propose DSANet, which disentangles camera characteristic information and extracts discriminative ID-related representation. TLM and FWG obtain features that are robust against temporal appearance destruction by cooperatively utilizing spatial and temporal information. DSANet can cope with various scenarios by augmenting the features of a new pair through SAO and performing auxiliary tasks. Experimental results demonstrate that our method is lightweight. Specifically, it achieves the highest performance on the LS-VID. Diverse visualization results illustrate qualitatively that the feature creation and disentanglement were as intended. We expect that the disentangling structure of our DSANet and SAO with auxiliary tasks will be used harmoniously with other methods in the future.


\noindent \footnotesize \textbf{Acknowledgement} This work was supported by the Institute of Information \& communications Technology Planning \& Evaluation(IITP) grant funded by the Korea government(MSIT) (No. 2021-0-00172, The development of human Re-identification and masked face recognition based on CCTV camera), the KIST Institutional Program(Project No.2E31051-21-203) and the Yonsei University Research Fund of 2021 (2021-22-0001). 
{\small
\bibliographystyle{ieee_fullname}
\bibliography{egbib}

\begin{thebibliography}{10}\itemsep=-1pt

\bibitem{chen2020temporal}
Guangyi Chen, Yongming Rao, Jiwen Lu, and Jie Zhou.
\newblock Temporal coherence or temporal motion: Which is more critical for
  video-based person re-identification?
\newblock In {\em European Conference on Computer Vision}, pages 660--676.
  Springer, 2020.

\bibitem{chen2020simple}
Ting Chen, Simon Kornblith, Mohammad Norouzi, and Geoffrey Hinton.
\newblock A simple framework for contrastive learning of visual
  representations.
\newblock In {\em International conference on machine learning}, pages
  1597--1607. PMLR, 2020.

\bibitem{chi2020pedhunter}
Cheng Chi, Shifeng Zhang, Junliang Xing, Zhen Lei, Stan~Z Li, and Xudong Zou.
\newblock Pedhunter: Occlusion robust pedestrian detector in crowded scenes.
\newblock In {\em Proceedings of the AAAI Conference on Artificial
  Intelligence}, volume~34, pages 10639--10646, 2020.

\bibitem{chu2021learning}
Sanghyeok Chu, Dongwan Kim, and Bohyung Han.
\newblock Learning debiased and disentangled representations for semantic
  segmentation.
\newblock {\em Advances in Neural Information Processing Systems},
  34:8355--8366, 2021.

\bibitem{deng2009imagenet}
Jia Deng, Wei Dong, Richard Socher, Li-Jia Li, Kai Li, and Li Fei-Fei.
\newblock Imagenet: A large-scale hierarchical image database.
\newblock In {\em 2009 IEEE conference on computer vision and pattern
  recognition}, pages 248--255. Ieee, 2009.

\bibitem{eom2021video}
Chanho Eom, Geon Lee, Junghyup Lee, and Bumsub Ham.
\newblock Video-based person re-identification with spatial and temporal memory
  networks.
\newblock In {\em Proceedings of the IEEE/CVF International Conference on
  Computer Vision}, pages 12036--12045, 2021.

\bibitem{fu2019sta}
Yang Fu, Xiaoyang Wang, Yunchao Wei, and Thomas Huang.
\newblock Sta: Spatial-temporal attention for large-scale video-based person
  re-identification.
\newblock In {\em Proceedings of the AAAI conference on artificial
  intelligence}, volume~33, pages 8287--8294, 2019.

\bibitem{gu2020appearance}
Xinqian Gu, Hong Chang, Bingpeng Ma, Hongkai Zhang, and Xilin Chen.
\newblock Appearance-preserving 3d convolution for video-based person
  re-identification.
\newblock In {\em European Conference on Computer Vision}, pages 228--243.
  Springer, 2020.

\bibitem{he2016deep}
Kaiming He, Xiangyu Zhang, Shaoqing Ren, and Jian Sun.
\newblock Deep residual learning for image recognition.
\newblock In {\em Proceedings of the IEEE conference on computer vision and
  pattern recognition}, pages 770--778, 2016.

\bibitem{he2020guided}
Lingxiao He and Wu Liu.
\newblock Guided saliency feature learning for person re-identification in
  crowded scenes.
\newblock In {\em European Conference on Computer Vision}, pages 357--373.
  Springer, 2020.

\bibitem{he2019foreground}
Lingxiao He, Yinggang Wang, Wu Liu, He Zhao, Zhenan Sun, and Jiashi Feng.
\newblock Foreground-aware pyramid reconstruction for alignment-free occluded
  person re-identification.
\newblock In {\em Proceedings of the IEEE/CVF international conference on
  computer vision}, pages 8450--8459, 2019.

\bibitem{hermans2017defense}
Alexander Hermans, Lucas Beyer, and Bastian Leibe.
\newblock In defense of the triplet loss for person re-identification.
\newblock {\em arXiv preprint arXiv:1703.07737}, 2017.

\bibitem{hou2021bicnet}
Ruibing Hou, Hong Chang, Bingpeng Ma, Rui Huang, and Shiguang Shan.
\newblock Bicnet-tks: Learning efficient spatial-temporal representation for
  video person re-identification.
\newblock In {\em Proceedings of the IEEE/CVF Conference on Computer Vision and
  Pattern Recognition}, pages 2014--2023, 2021.

\bibitem{hou2020temporal}
Ruibing Hou, Hong Chang, Bingpeng Ma, Shiguang Shan, and Xilin Chen.
\newblock Temporal complementary learning for video person re-identification.
\newblock In {\em European conference on computer vision}, pages 388--405.
  Springer, 2020.

\bibitem{hou2019vrstc}
Ruibing Hou, Bingpeng Ma, Hong Chang, Xinqian Gu, Shiguang Shan, and Xilin
  Chen.
\newblock Vrstc: Occlusion-free video person re-identification.
\newblock In {\em Proceedings of the IEEE/CVF conference on computer vision and
  pattern recognition}, pages 7183--7192, 2019.

\bibitem{huang2018multimodal}
Xun Huang, Ming-Yu Liu, Serge Belongie, and Jan Kautz.
\newblock Multimodal unsupervised image-to-image translation.
\newblock In {\em Proceedings of the European conference on computer vision
  (ECCV)}, pages 172--189, 2018.

\bibitem{jiang2021ssn3d}
Xiaoke Jiang, Yu Qiao, Junjie Yan, Qichen Li, Wanrong Zheng, and Dapeng Chen.
\newblock Ssn3d: Self-separated network to align parts for 3d convolution in
  video person re-identification.
\newblock In {\em Proceedings of the AAAI Conference on Artificial
  Intelligence}, volume~35, pages 1691--1699, 2021.

\bibitem{kim2022occluded}
Minjung Kim, MyeongAh Cho, Heansung Lee, Suhwan Cho, and Sangyoun Lee.
\newblock Occluded person re-identification via relational adaptive feature
  correction learning.
\newblock In {\em ICASSP 2022-2022 IEEE International Conference on Acoustics,
  Speech and Signal Processing (ICASSP)}, pages 2719--2723. IEEE, 2022.

\bibitem{lee2021learning}
Jungsoo Lee, Eungyeup Kim, Juyoung Lee, Jihyeon Lee, and Jaegul Choo.
\newblock Learning debiased representation via disentangled feature
  augmentation.
\newblock {\em Advances in Neural Information Processing Systems},
  34:25123--25133, 2021.

\bibitem{lee2021dranet}
Seunghun Lee, Sunghyun Cho, and Sunghoon Im.
\newblock Dranet: Disentangling representation and adaptation networks for
  unsupervised cross-domain adaptation.
\newblock In {\em Proceedings of the IEEE/CVF conference on computer vision and
  pattern recognition}, pages 15252--15261, 2021.

\bibitem{li2019global}
Jianing Li, Jingdong Wang, Qi Tian, Wen Gao, and Shiliang Zhang.
\newblock Global-local temporal representations for video person
  re-identification.
\newblock In {\em Proceedings of the IEEE/CVF international conference on
  computer vision}, pages 3958--3967, 2019.

\bibitem{li2019multi}
Jianing Li, Shiliang Zhang, and Tiejun Huang.
\newblock Multi-scale 3d convolution network for video based person
  re-identification.
\newblock In {\em Proceedings of the AAAI Conference on Artificial
  Intelligence}, volume~33, pages 8618--8625, 2019.

\bibitem{li2018diversity}
Shuang Li, Slawomir Bak, Peter Carr, and Xiaogang Wang.
\newblock Diversity regularized spatiotemporal attention for video-based person
  re-identification.
\newblock In {\em Proceedings of the IEEE conference on computer vision and
  pattern recognition}, pages 369--378, 2018.

\bibitem{li2021diverse}
Yulin Li, Jianfeng He, Tianzhu Zhang, Xiang Liu, Yongdong Zhang, and Feng Wu.
\newblock Diverse part discovery: Occluded person re-identification with
  part-aware transformer.
\newblock In {\em Proceedings of the IEEE/CVF Conference on Computer Vision and
  Pattern Recognition}, pages 2898--2907, 2021.

\bibitem{luo2019bag}
Hao Luo, Youzhi Gu, Xingyu Liao, Shenqi Lai, and Wei Jiang.
\newblock Bag of tricks and a strong baseline for deep person
  re-identification.
\newblock In {\em Proceedings of the IEEE/CVF conference on computer vision and
  pattern recognition workshops}, pages 0--0, 2019.

\bibitem{mclaughlin2016recurrent}
Niall McLaughlin, Jesus~Martinez Del~Rincon, and Paul Miller.
\newblock Recurrent convolutional network for video-based person
  re-identification.
\newblock In {\em Proceedings of the IEEE conference on computer vision and
  pattern recognition}, pages 1325--1334, 2016.

\bibitem{van2008visualizing}
Laurens Van~der Maaten and Geoffrey Hinton.
\newblock Visualizing data using t-sne.
\newblock {\em Journal of machine learning research}, 9(11), 2008.

\bibitem{wang2021residual}
Yifei Wang, Zhengyang Geng, Feng Jiang, Chuming Li, Yisen Wang, Jiansheng Yang,
  and Zhouchen Lin.
\newblock Residual relaxation for multi-view representation learning.
\newblock {\em Advances in Neural Information Processing Systems},
  34:12104--12115, 2021.

\bibitem{wang2021pyramid}
Yingquan Wang, Pingping Zhang, Shang Gao, Xia Geng, Hu Lu, and Dong Wang.
\newblock Pyramid spatial-temporal aggregation for video-based person
  re-identification.
\newblock In {\em Proceedings of the IEEE/CVF International Conference on
  Computer Vision}, pages 12026--12035, 2021.

\bibitem{woo2018cbam}
Sanghyun Woo, Jongchan Park, Joon-Young Lee, and In~So Kweon.
\newblock Cbam: Convolutional block attention module.
\newblock In {\em Proceedings of the European conference on computer vision
  (ECCV)}, pages 3--19, 2018.

\bibitem{wu2018exploit}
Yu Wu, Yutian Lin, Xuanyi Dong, Yan Yan, Wanli Ouyang, and Yi Yang.
\newblock Exploit the unknown gradually: One-shot video-based person
  re-identification by stepwise learning.
\newblock In {\em Proceedings of the IEEE conference on computer vision and
  pattern recognition}, pages 5177--5186, 2018.

\bibitem{ye2021deep}
Mang Ye, Jianbing Shen, Gaojie Lin, Tao Xiang, Ling Shao, and Steven~CH Hoi.
\newblock Deep learning for person re-identification: A survey and outlook.
\newblock {\em IEEE transactions on pattern analysis and machine intelligence},
  44(6):2872--2893, 2021.

\bibitem{zhang2020multi}
Zhizheng Zhang, Cuiling Lan, Wenjun Zeng, and Zhibo Chen.
\newblock Multi-granularity reference-aided attentive feature aggregation for
  video-based person re-identification.
\newblock In {\em Proceedings of the IEEE/CVF conference on computer vision and
  pattern recognition}, pages 10407--10416, 2020.

\bibitem{zhang2020relation}
Zhizheng Zhang, Cuiling Lan, Wenjun Zeng, Xin Jin, and Zhibo Chen.
\newblock Relation-aware global attention for person re-identification.
\newblock In {\em Proceedings of the ieee/cvf conference on computer vision and
  pattern recognition}, pages 3186--3195, 2020.

\bibitem{zheng2016mars}
Liang Zheng, Zhi Bie, Yifan Sun, Jingdong Wang, Chi Su, Shengjin Wang, and Qi
  Tian.
\newblock Mars: A video benchmark for large-scale person re-identification.
\newblock In {\em European conference on computer vision}, pages 868--884.
  Springer, 2016.

\bibitem{zheng2019joint}
Zhedong Zheng, Xiaodong Yang, Zhiding Yu, Liang Zheng, Yi Yang, and Jan Kautz.
\newblock Joint discriminative and generative learning for person
  re-identification.
\newblock In {\em proceedings of the IEEE/CVF conference on computer vision and
  pattern recognition}, pages 2138--2147, 2019.

\bibitem{zhong2020random}
Zhun Zhong, Liang Zheng, Guoliang Kang, Shaozi Li, and Yi Yang.
\newblock Random erasing data augmentation.
\newblock In {\em Proceedings of the AAAI conference on artificial
  intelligence}, volume~34, pages 13001--13008, 2020.

\bibitem{zhou2017see}
Zhen Zhou, Yan Huang, Wei Wang, Liang Wang, and Tieniu Tan.
\newblock See the forest for the trees: Joint spatial and temporal recurrent
  neural networks for video-based person re-identification.
\newblock In {\em Proceedings of the IEEE conference on computer vision and
  pattern recognition}, pages 4747--4756, 2017.

\bibitem{zhu2020identity}
Kuan Zhu, Haiyun Guo, Zhiwei Liu, Ming Tang, and Jinqiao Wang.
\newblock Identity-guided human semantic parsing for person re-identification.
\newblock In {\em European Conference on Computer Vision}, pages 346--363.
  Springer, 2020.

\bibitem{zou2020joint}
Yang Zou, Xiaodong Yang, Zhiding Yu, BVK Kumar, and Jan Kautz.
\newblock Joint disentangling and adaptation for cross-domain person
  re-identification.
\newblock In {\em European Conference on Computer Vision}, pages 87--104.
  Springer, 2020.

\end{thebibliography}
}

\end{document}